\documentclass[final]{elsarticle}
\usepackage{hyperref}
\usepackage{verbatim}

\usepackage{amssymb}% http://ctan.org/pkg/amssymb
\usepackage{pifont}% http://ctan.org/pkg/pifont
\newcommand{\cmark}{\ding{51}}%
\newcommand{\xmark}{\ding{55}}%

\begin{document}

\begin{frontmatter}

\title{Understanding in Artificial Intelligence}

\author[mymainaddress]{Stefan Maetschke}
\author[mymainaddress]{David Martinez Iraola}
\author[mymainaddress]{Pieter Barnard}
\author[mymainaddress]{Elaheh ShafieiBavani}
\author[mymainaddress]{Peter Zhong}
\author[mymainaddress]{Ying Xu}
\author[mymainaddress]{Antonio Jimeno Yepes}

\address[mymainaddress]{IBM Research Australia, Melbourne, VIC, AUS}

% -----------------------------------------------------------------------------------------------------------------------
\begin{abstract}
Current Artificial Intelligence (AI) methods, most based on deep learning, have facilitated progress in several fields, including computer vision and natural language understanding.
The progress of these AI methods is measured using benchmarks designed to solve challenging tasks, such as visual question answering.
A question remains of how much understanding is leveraged by these methods and how appropriate are the current benchmarks to measure understanding capabilities.
To answer these questions, we have analysed existing benchmarks and their understanding capabilities, defined by a set of understanding capabilities, and current research streams.
We show how progress has been made in benchmark development to measure understanding capabilities of AI methods and we review as well how current methods develop understanding capabilities.

\end{abstract}

% -----------------------------------------------------------------------------------------------------------------------

\end{frontmatter}

%\linenumbers

% -----------------------------------------------------------------------------------------------------------------------
\section{Introduction}

Recent advancements in deep learning have facilitated tremendous progress
in computer vision, speech processing, natural language understanding
and many other domains~\citep{lecun2015deep,schmidhuber2015deep}. However, this progress is largely driven by
increased computational power, namely GPU's, and bigger data sets but
not due to radically new algorithms or knowledge
representations. Artificial Neural Networks and Stochastic Gradient
Descent, popularized in the 80's~\citep{rummelhart1986learning}, remain
the fundamental building blocks for most modern AI systems.

While very successful for many applications, especially in vision, the
purely deep-learning based approach has significant weaknesses. For
instance, CNN's struggle with~same-different relations
\citep{ricci2018same}, fail when long-chained reasoning is needed
\citep{johnson2017clevr}, are non-decomposable, cannot easily
incorporate symbolic knowledge, and are hampered by a lack of
model interpretability. Many current methods essentially compute higher order
statistics over basic elements such as pixels, phonemes, letters or
words to process inputs but do not explicitly model the building
blocks and their relations in a (de)composable and interpretable
way. 

On the other hand, there is a long tradition of symbolic knowledge
representation, formal logic and reasoning. However, such approaches,
struggle with contradicting, incomplete or fuzzy information.

%Consequently, existing approaches are unable to solve many tasks
%that are easily performed by humans.

In this article, we focus on presenting existing benchmarks and
research streams that go beyond purely deep-learning or symbolic
methods, and describe how they represent and apply knowledge in a way that aims at
human-level understanding. These works explore knowledge
representations that are hierarchically structured, compositional, and
unify different modalities. We will start by providing background on
the definition of "understanding" and its implications for
artificial intelligence in Section~\ref{sec:understanding}. As we will discuss, a way
of evaluating understanding capabilities is by using benchmarks to
measure performance against other systems or humans. We will present
and analyse existing benchmarks in Section~\ref{sec:benchmarks}. Once we have a
definition and benchmarks to measure understanding, we will introduce
the different research streams recently developed in this area (Section~\ref{sec:review}). Finally, we discuss and summarise our study in Sections~\ref{sec:discussion} and \ref{sec:conclusion}.

\section{Definition of "understanding"}
\label{sec:understanding}

The Cambridge Dictionary defines "understanding" as \textit{to know why or how something happens or works}.
Similar definitions are available in fields such as psychology~\citep{bereiter2005education}, philosophy~\citep{kant1998critique} and computer science~\citep{chaitin2006limits}\footnote{More specifically in algorithmic information theory an argument is made to relate understanding to data compression.}.

%%%% Desirable properties for a system that understands
Understanding has been recognized as a relevant component in artificial intelligence~\citep{lake2017building} and building systems with understanding capabilities is a significant step towards artificial general intelligence.
There are several theories regarding the inner workings of human intelligence and how an artificial intelligence could be built~\citep{chollet2019measure}.

Current state-of-the-art methods implementing artificial intelligence -- sometimes even surpassing human performance -- still lack an understanding of the tasks they are applied to.
For instance, systems that learn to play Atari games~\citep{mnih2013playing} or AlphaGO~\citep{silver2017mastering} require a large amount of example games to tune the underlying game model. However, even after extensive training no general understanding of the game is generated and the model is not reusable for other, even closely related, tasks.

Another example is in image analytics, where deep neural networks are currently the state of the art for many tasks, for instance in medical image diagnosis, face recognition and self-driving cars.
So called "adversarial attacks", in which images are perturbed by tiny amounts that are essentially invisible to humans, can cause these networks to wrongly classify objects with high confidence~\citep{nguyen2015deep}. This shows that these networks have no actual understanding of the scene but rely on higher-order pixel distributions for classification.
Similarly state-of-the-art natural language understanding systems are sensitive to small changes in the input text that humans are not affected by~\citep{jin2019bert}.

A formal definition and evaluation of a system's understanding capabilities is challenging~\citep{chollet2019measure}.
Similar to evaluating intelligence, "understanding" could be measured by human evaluators who judge the output of a system in a similar way as a Turing test has been proposed to recognize intelligence. However, such a measurement is labor intensive and subjective.

An alternative way of evaluating the understanding capabilities of a system is by benchmarking its performance for specific data sets and tasks.
Using benchmarks is a common practice to evaluate machine learning systems and allow for a reproducible and objective evaluation.

Benchmarks with increasingly more complex tasks can be defined to ensure that algorithms solves increasingly more complex problems~\citep{santoro2018measuring,hernandez2020ai,crosby2020building}, with the hope to move closer and closer to a system that shows truly intelligent behavior.

Similar to the assessment of human intelligence, there is the need to prevent memorization, in which success in a benchmark might be achieved by memorizing questions and answers. Also biases in the benchmark data set need to be considered~\citep{goyal2017making}, since they can be exploited; impeding the evaluation of understanding capabilities.

Examples of recent benchmarks such as CLEVR~\citep{johnson2017clevr} overcome these problems by generating synthetic data with strictly controlled distributions to avoid any biases.
However, the understanding capabilities on synthetic data are often limited,
since the complexity in comparison with real-world scenario is severely reduced.
For instance, the CLEVR benchmark requires the recognition and localization of a very small number of objects with a small set of fixed properties. It has successfully been solved using neuro-symbolic approaches, surpassing human performance~\citep{mao2019neuro}.
To move towards more complex understanding capabilities, requires the creation and use of benchmarks that exhibit an increasing level of complexity.

Another property relevant to understanding is compositionality ~\citep{lake2017building,lake2015human}, which has been discussed in detail for vision tasks~\citep{biederman1987recognition} and natural language understanding~\citep{manning2016understanding}. It essentially refers to a system's capability to identify the composing parts of an object or a problem and being able to reuse parts in new combinations to solve related but different tasks more efficiently, e.g. with less data or faster.

As the level of understanding of the system improves, we will consider the integration of existing information in ontologies and knowledge graphs to quickly and easily increase the systems knowledge.
Being able to integrate knowledge directly could support the transfer of skills across tasks or domains, which would contribute to the adaptability of AI that understand to new problems.

\section{Benchmarks}
\label{sec:benchmarks}
%% Use the categorization proposed in the white paper

As mentioned in section~\ref{sec:understanding}, benchmarks play an important role to evaluate the capabilities of existing AI systems.
Therefore, they can be seen as guiding the development of algorithm's to allow more complex tasks to be solved. In this section, recent advances in benchmarks are explored in each of the categories image analytics (section~\ref{sub: Image analytics benchmarks}), natural language processing (section~\ref{sub: Natural language processing benchmarks}), visual question answering (section~\ref{sec:vqa benchmarks}), common sense inference (section~\ref{sub: Commonsense inference benchmarks}) and common sense that require NLP (section ~\ref{sub: NLP common sense benchmarks}).

%These problems are chosen or designed with a focus on the required capabilities of an AI that understands:

%\begin{itemize}
%	\item CAP1: hierarchical and compositional knowledge representation,
%	\item CAP2: multi-modal structure-to-structure mapping,
%	\item CAP3: integrates symbolic and non-symbolic knowledge,
%	\item CAP4: supports symbolic reasoning with uncertainties.
%\end{itemize}

\subsection{Image analytics benchmarks}
\label{sub: Image analytics benchmarks}

Current state-of-the-art methods in processing images and text are evaluated on benchmarks that require some level of understanding.
Arguably, for many of these benchmarks the level of understanding is limited and neural network based function approximators are able to provide a high level of accuracy. The following examples show existing benchmarks on a multiple of data modalities that either are, or have been considered for state-of-the-art model evaluation.

Image analytics has seen a revolution with deep learning and many data sets have seen their performance improved to become similar to human performance in image classification (such as ImageNet~\cite{deng2009imagenet}) and some object detection benchmarks. In object detection, the task is to identify objects from a predefined set of images. Annotated sets are used to identify these objects. Example data sets include, COCO (Common Objects in Context)~\cite{lin2014microsoft}~\footnote{\url{https://cocodataset.org/index.ht}} which contains over 330K images and 1.5 million annotations, which can be used as well for image segmentation. The SHAPE benchmark~\cite{shilane2004princeton} offers a set of objects in 3D to be identified.

\subsection{Natural language processing (NLP) benchmarks}
\label{sub: Natural language processing benchmarks}

A large portion of knowledge is available in unstructured text format, thus providing methods to understand language are relevant. Even if performance in natural language processing has been lagging behind the use of deep learning, the field is catching up with current developments. Some tasks include natural language understanding, as provided by the Glue (General Language Understanding Evaluation) benchmark~\cite{wang2018glue} that contains several tasks for natural language understanding.

Question answering is found in many common and popular tasks, in which the objective is to find the answer to a question given in text. Even if we can see that the task might require limited understanding capabilities in most cases, these tasks have motivated research into more challenging benchmarks. Examples of benchmarks for question answering include SQUAD~\cite{rajpurkar2016squad}, with over 100K questions and answers crowdsourced using Wikipedia articles, this benchmark has provided a valuable resource to test novel NLP methods.
Question answering has taken different shapes and forms with increasing level of complexity.

%NarrativeQA~\cite{kovcisky2018narrativeqa}: provides full novels and other long texts as evidence documents and contains approximately 30 crowdsourced questions per text.

%TriviaQA~\cite{joshi2017triviaqa}: a corpus of webcrawled trivia and quiz-league websites together with evidence documents from the web.

%HotpotQA~\cite{yang2018hotpotqa} and WorldTree~\cite{jansen2018worldtree}: to provide explicit gold explanations that serve as training and evaluation instruments for multi-hop inference models. These data sets were created to overcome the existing limits on the length of inferences in traversing the knowledge graphs~\cite{khashabi2019capabilities}.

%Domain specific benchmarks and competitions are available as well, e.g. BioASQ~\cite{tsatsaronis2015overview}.

The AI2 Reasoning Challenge (ARC)~\url{https://allenai.org/data/arc} is a dataset of 7,787 genuine grade-school level, multiple-choice science questions, assembled to encourage research in advanced question-answering.

In contrast to these question and answering bechmarks, other tasks have appeared that require additional attention to the context. Examples of those benchmarks include LAMBADA (LAnguage Modeling Broadened to Account for Discourse Aspects)~\cite{paperno2016lambada}, in which it is required to have looked at the whole text before an answer can be provided. More examples of novel question and answering tasks have appeared such as BREAK~\cite{Wolfson2020Break}, which is a new question understanding benchmark data set that combines multiple older VQA data sets. In addition the data set provides high-quality, human-generated question decompositions, so called "Question Decomposition Meaning Representations" (QDMR) that are similar to the program traces of the CLEVR data set \citep{johnson2017clevr} but at a slightly higher and richer level of abstraction. The data set is composed of 36k question and answer pairs with the corresponding question decompositions.

%\begin{figure}
%	\centering
%	\includegraphics[width=0.7\textwidth]{pics/break}
%	\caption{Examples of question decompositions for different modalities in BREAK. Image adopted from \citep{Wolfson2020Break}.}
%\label{fig:break}
%\end{figure}

Furthermore, the creators of the data set also conduct a \href{https://allenai.github.io/Break/#leaderboard}{Challenge}. The data set and challenge offers an excellent opportunity to evaluate current state-of-the-art methods and to identify the issues to overcome for understanding systems.

\subsection{Visual Question Answering (VQA) benchmarks}
\label{sec:vqa benchmarks}

In addition to benchmarks covering only one data modality, e.g. text or images, there is recent work that combines both modalities in what is called Visual Question Answering (VQA).
Examples of those data sets include VQAv1~\cite{agrawal2017c},  VQAv2~\cite{goyal2017making}, and Visual Genome~\cite{antol2015vqa}.

%%Link or move the next two paragraphs to benchmarks
An analysis of VQA algorithms by \cite{kafle2017analysis} found that
VQA data sets with natural images are biased. For instance, in many
cases questions regarding color tend to be much more frequent than
others and VQA can exploit this bias, circumventing the need for true
scene understanding. Consequently, the authors created the \emph{Task
	Driven Image Understanding Challenge} (TDIUC) data set to reduce
this bias. Another common balanced data set is the VQAv2 data set by
\cite{goyal2017making}, where every question is associated with a
pair of similar images but different answers.

Similarly, synthetic data sets such as
SHAPES~\citep{Andreas2015DeepCQ} and CLEVR~\citep{johnson2017clevr}
were created to control bias but also to determine which question
types pose problems for VQA algorithms; for instance, counting of
objects~\citep{hu2017learning}. SHAPES is a small data set with 64
images of colored geometric shapes in different arrangements. CLEVR is
much larger with 999,968 questions, 13 question types and 100,000
images of 3D rendered geometric objects such as cubes, spheres and
cylinders of two different sizes, eight colors and two materials.

To overcome the bias that exist in previous benchmarks, CLEVR followed the approach of generating synthetic data, which provides control over data being generated and at the same time provides challenging tasks that require reasoning about the object in the scene.
Neuro-symbolic reasoning methods~\cite{gan2017vqs} have surpassed human performance on this task.
Recently, this benchmark has been extended with an animated version in which the task is to predict a future outcome of the animation~\cite{yi2019clevrer}.

% Combines images and text, this is visual question answering.
Additional benchmarks have been generated which address more complex reasoning such as Math SemEval~\cite{hopkins-etal-2019-semeval} in which the task is to solve mathematical problems which in some cases require understanding of the image in which the problem is depicted.

\subsection{Commonsense inference benchmarks}
\label{sub: Commonsense inference benchmarks}
%% Ella
In this area of AI, there is a critical need for integrating different modes of reasoning (e.g. symbolic reasoning through deduction and statistical reasoning based on large amount of data), as well as benchmarks and evaluation metrics that can quantitatively measure research progress ~\cite{davis2015commonsense}.

In recent years, there has been a surge of research activities in the NLP community to tackle commonsense reasoning and inference through ever-growing benchmark tasks. These tasks range from earlier textual entailment tasks, e.g. the Recognizing Textual Entailment (RTE) Challenges~\citep{dagan2005pascal}, to more recent
tasks that require a comprehensive understanding of everyday physical and social commonsense, e.g. the Story Cloze Test~\citep{mostafazadeh2016corpus} or SWAG~\citep{zellers2018swag}. An increasing effort has been
devoted to extracting commonsense knowledge from existing data (e.g. Wikipedia) or acquiring it directly from crowd workers. Many learning and inference approaches have been developed for these benchmark tasks which range from earlier symbolic and statistical approaches to more recent neural approaches~\cite{storks2019commonsense}. 

Many commonsense benchmarks are based upon classic language processing problems. The scope of these benchmark tasks ranges from more focused tasks, such as coreference resolution and named entity recognition, to more comprehensive tasks and applications, such as question answering and textual entailment. More focused tasks tend to be useful in creating component technology for NLP systems, building upon each other toward more comprehensive tasks. Meanwhile, rather than restricting tasks by the types of language processing skills required to perform them, a common characteristic of earlier benchmarks, recent benchmarks are more commonly geared toward particular types of commonsense knowledge and reasoning. Some benchmark tasks focus on singular commonsense reasoning processes, e.g., temporal reasoning, requiring a small amount of commonsense knowledge, while others focus on entire domains of knowledge, e.g., social psychology, thus requiring a larger set of related reasoning skills. Furthermore, some benchmarks include a more comprehensive mixture of everyday commonsense knowledge, demanding a more complete commonsense reasoning skill set~\citep{storks2019commonsense}. Some of the commonsense inference benchmarks in NLP are as follow:

\begin{itemize}
	\item TriviaQA \citep{joshi2017triviaqa}: a corpus of webcrawled trivia and quiz-league websites together with evidence documents from the web.
	\item CommonsenseQA \citep{talmor2018commonsenseqa}: consists of 9,000 crowdsourced multiple-choice questions with a focus on relations between entities that appear in ConceptNet.
	\item NarrativeQA \citep{kovcisky2018narrativeqa}: provides full novels and other long texts as evidence documents and contains approximately 30 crowdsourced questions per text.
	\item NewsQA \citep{trischler2016newsqa}: provides news texts with crowdsourced questions and answers, which are spans of the evidence documents.
	\item The Story cloze test and the ROC data set \citep{mostafazadeh2016corpus}: systems have to find the correct ending to a 5-sentence story, using different types of commonsense knowledge.
	\item SWAG \citep{zellers2018swag}: is a Natural Language Inference (NLI) data set of 113k highly varied grounded situations for commonsense application with a focus on difficult commonsense inferences.
	\item WikiSQL \citep{zhong2017seq2sql}: is a corpus of 87,726 hand-annotated instances of natural language questions, SQL queries, and SQL tables. This data set was created as the benchmark data set for the table-based question answering task.
	\item Datasets for graph-based learning \citep{xu2018graph2seq}: i) SDP\textsubscript{DAG} whose graphs are directed acyclic graphs (DAGs); ii) SDP\textsubscript{DCG} whose graphs are directed cyclic graphs
	(DCGs) that always contain cycles; iii) SDP\textsubscript{SEQ} whose graphs are essentially sequential lines.
	\item HotpotQA \citep{yang2018hotpotqa} and WorldTree \citep{jansen2018worldtree}: to provide explicit gold explanations that serve as training and evaluation instruments for multi-hop inference models. These data sets were created to overcome the existing limits on the length of inferences in traversing the knowledge graphs \citep{khashabi2019capabilities}.
	\item Event2Mind \citep{rashkin2018event2mind}: this corpus
	has 25,000 narrations about everyday activities and situations, for which the best performing model is ConvNet \citep{rashkin2018event2mind}.
	\item Winograd and Winograd NLI schema Challenge \citep{mahajan2018winograd}: Employs Winograd Schema questions that require the resolution of anaphora i.e. the model should identify the antecedent of an ambiguous pronoun.
	
\end{itemize}

Common sense knowledge bases include:

\begin{itemize}
	\item ConceptNet \citep{speer2017conceptnet}: contains over 21 million edges and 8 million nodes (1.5 million nodes in the partition for the English vocabulary), generating triples of the form ($C1$, $r$, $C2$): the natural-language concepts $C1$ and $C2$ are associated by commonsense relation $r$.
	\item WebChild \citep{tandon2017webchild}: is a large collection of commonsense knowledge, automatically extracted from Web contents. WebChild contains triples that connect nouns with adjectives via fine-grained relations. The arguments of these assertions, nouns and adjectives, are disambiguated by mapping them onto their proper WordNet senses.
	\item Never Ending Language Learner (NELL) \citep{mitchell2018never}: is C.M.U.’s learning agent that actively learns relations from the web and keeps expanding its knowledge base 24/7 since 2010. It has about 80 million facts from the web with varying confidence. It continuously learns facts and also keeps improving its reading competence and thus learning accuracy.
	\item ATOMIC \citep{sap2019atomic} is a new knowledge-base that focuses on procedural knowledge. Triples are of the form (Event, $r$, \{Effect|Persona|Mental-state\}), where head and tail are short sentences or verb phrases and $r$ represents an if-then relation type.
	
\end{itemize}

In terms of tackling the task of commonsense inference in NLP, some methods propose to explicitly incorporate commonsense knowledge, and more recent works focus on using pre-trained deep learning representations such as BERT or XLNet.  \cite{da2019cracking} investigated a number of aspects of BERT's commonsense representation abilities to provide a better understanding of the capability of these models. Their findings demonstrate that BERT is able to encode various commonsense features in its embedding space; and if there is deficiency in its representation, the issue can be solved by pre-training with additional data related to the deficient attributes. \cite{li-etal-2019-pingan} follow a similar approach over the COIN 2019 shared task data, and they obtain the best performance for both tasks by relying on XLNet to get contextual representations.

On the other hand, explicitly incorporating commonsense requires injecting knowledge into pre-trained deep learning models. Inspired by \cite{bauer2018commonsense}, \cite{ma2019towards} proposed to convert concept-relation tokens into regular tokens, which are later used to generate a pseudo-sentence. Embeddings of the converted pseudo-sentences are concatenated with other representations for commonsense inference.

Apart from reading comprehension, there are other commonsense inference tasks that are being used for evaluation. In the work of \cite{huminski2019commonsense} the goal is to perform commonsense inference in human-robot communication. This is an important task because human-human communications often describe the expected change of state after an action, and the actions to achieve the change are not detailed. In their work a method to transform high-level result-verb commands into action-verb commands is implemented and evaluated. Because of the lack of manually annotated data, the proposed solution is a pipeline relying on WordNet, ngram frequencies, and a search engine to predict the mapping.

Another relevant commonsense knowledge task is the prediction of event sequences. This has been addressed by relying on sequence-to-sequence models trained on annotated data, and data sets have been manually annotated for implementation and evaluation of these systems~\citep{nguyen-etal-2017-sequence}. Recent work on this task includes the use of conditional variational autoencoders for improved diversity, and the extension of original annotated data to cover multiple follow-up events~\citep{kiyomaru-etal-2019-diversity}.

Semantic plausibility is another commonsense-related task that has been used as a test bed for analyzing knowledge representations. In \citep{porada-etal-2019-gorilla}, self-supervision is applied to learn from text via pre-trained models (BERT) and state-of-the-art performance is achieved without requiring manual annotations.

Finally, recent research in this area has focused on physical properties of objects, and being able to make comparisons by applying commonsense reasoning. As an example of this task, \cite{goel-etal-2019-pre} observe that probing pre-trained models (GloVe, ELMo, and BERT) can reliably perform physical comparisons between objects.

In terms of graph-based commonsense inference, question answering in bAbI~\citep{li2015gated}, Shortest path, and Natural Language Generation \citep{song2017amr} are the tasks that have been modelled as a Sequence to Sequence learning (Seq2Seq) technique~\citep{xu2018graph2seq} to tackle the challenge of achieving accurate conversion from graph to the appropriate sequence. To this end, a general end-to-end graph-to-sequence neural encoder-decoder model is proposed for mapping an input graph to a sequence of vectors using an attention-based LSTM method to decode the target sequence from these vectors~\citep{xu2018graph2seq}. Using the proposed bi-directional node embedding aggregation strategy, the model converges rapidly to the optimal performance and outperforms existing graph neural networks, Seq2Seq, and Tree2Seq models.

Some graph-based approaches take a step further and try to answer complex questions requiring a combination of multiple facts, and identify why those answers are correct~\citep{thiem2019extracting}. Combining multiple facts to answer questions is often modeled as a \textit{multi-hop} graph traversal problem, which suffers from semantic drift, or the tendency for chains of reasoning to \textit{drift} to unrelated topics, and this semantic drift greatly limits the number of facts that can be combined in both free text or knowledge base inference. This issue is addressed by extracting large high-confidence multi-hop inference patterns, generated by abstracting large-scale explanatory structure from a corpus of detailed explanations. Given that, a prototype tool for identifying common inference patterns from corpora of semi-structured explanations has been released.

%\subsection{Reasoning tasks} %% Revise: Antonio

%List of benchmarks:

%Task Driven Understanding Language (TDUIL)

%FlashFill benchmarks~\cite{gulwani2011automating}  it did not generalize well to larger
%programs nor does it seem to scale well computationally for more complex
%tasks.

%Satisfiability Modulo Theories (SMT) problems~\cite{ [De Moura and Bjrner, 2008] to learn
%SVRT concepts [Fleuret et al., 2011] such as spatial relationships between objects.

\subsection{Commonsense inference benchmarks that need Natural Language Processing}
\label{sub: NLP common sense benchmarks}

\begin{itemize}
	
	\item CommonsenseQA~\cite{talmor2018commonsenseqa} consists of 9,
	crowdsourced multiple-choice questions with a focus on relations
	between entities that appear in ConceptNet.
	
	\item NewsQA~\cite{trischler2016newsqa}: provides news texts with
	crowdsourced questions and answers, which are spans of the evidence
	documents.
	
	\item The Story cloze test and the ROC data set \cite{mostafazadeh2016corpus}
	systems have to find the correct ending to a 5-sentence story,
	using different types of commonsense knowledge.
	
	\item SWAG~\cite{zellers2018swag}: is a Natural Language Inference (NLI)
	data set of 113k highly varied grounded situations for commonsense
	application with a focus on difficult commonsense inferences.
	
	\item WikiSQL~\cite{zhong2017seq2sql}: is a corpus of 87,726 hand-annotated
	instances of natural language questions, SQL queries, and SQL
	tables. This data set was created as the benchmark data set for the
	table-based question answering task.
	
	\item Datasets for graph-based learning~\cite{xu2018graph2seq}: i) SDPDAG
	whose graphs are directed acyclic graphs (DAGs); ii) SDPDCG whose
	graphs are directed cyclic graphs (DCGs) that always contain cycles;
	iii) SDPSEQ whose graphs are essentially sequential lines.
	
	\item Event2Mind~\cite{rashkin2018event2mind}: this corpus has 25,000 narrations
	about everyday activities and situations, for which the best
	performing model is ConvNet.
	
	\item Winograd and Winograd NLI schema Challenge~\cite{mahajan2018winograd}:
	Employs Winograd Schema questions that require the resolution of
	anaphora i.e. the model should identify the antecedent of an
	ambiguous pronoun.
	
\end{itemize}

There has been progress in defining tasks for measuring understanding, and recently there have been new development in benchmark requiring and understanding (e.g. the Abstraction and Reasoning Corpus (ARC)~\cite{chollet2019measure}).
Several directions might be relevant to consider for extending the current approaches.
Such extensions would include understanding of the compositions of objects in different media and being able to reason about these compositions.
As well, integrate different media through a common representation and perform reasoning.

\section{Research streams}
\label{sec:review}

An AI that can understand requires certain technical capabilities. In this paper the following are considered as relevant features of such a system: (i) capable of hierarchical and compositional knowledge representation, (ii) provides multi-modal structure-to-structure mapping, (iii) integrates symbolic and non-symbolic knowledge, and (iv) supports symbolic reasoning with uncertainties.

Different research streams address these desired capabilities in some way, and this paper describes the main trends. Recently, there has been a renewed focus on combining the best of
neural networks and symbolic systems \citep{marcus2020decade}. These
so-called "neuro-symbolic" systems \citep{Besold2017NeuralSymbolicLA}
have achieved impressive improvements in Visual Question Answering
(VQA) tasks (CLEVR~\citep{johnson2017clevr}, Neuro-symbolic concept
learner~\citep{mao2019neuro}), even if they are currently limited to
image scenes with simple objects such as cubes, spheres and
cylinders\footnote{VQA systems have been applied to real-world
	scenes but even then objects are treated as units and not as
	entities composed of parts, e.g. a person with head, arms, legs and
	body. Furthermore, performance is significantly lower for real-world
	scenes. }. We start describing the use of neuro-symbolic systems in
Visual Question Answering in Section~\ref{sec:VQA}.

Another related research trend aims at hierarchically structured knowledge representations for different tasks (Section~\ref{sec:hierarchical_representations}), and in this stream we find methods that focus on multimodal representation (Section~\ref{sec:hierarchical_multi_modal}), tree-based representations (Section~\ref{sec:tree-lstm}), hyperbolic embeddings (Section~\ref{sec:hyperbolic_embedding}), and knowledge graph learning (Section~\ref{sec:knowledge graphs}).

Understanding images and documents is another of the main areas of research where deep understanding capabilities are required. Fully processing an image scene requires comprehension of the object it contains, and the relations between them. For documents, figures, tables and other structured images need to be
decomposed into elements to enable reasoning.  Relevant research
regarding scene and document understanding will be discussed in
section~\ref{sec:scene_graph}.

A required capability for understanding is to learn from example data a (partially) symbolic knowledge representation that allows reasoning and question answering. Research in \nameref{sec:program_synthesis} (section~\ref{sec:program_synthesis}),
\nameref{sub:neuro-symbolic} (section~\ref{sub:neuro-symbolic}) and \nameref{sec:common_sense_inference} (section~\ref{sec:common_sense_inference}) are of specific interest here.

\subsection{Visual Question Answering}
\label{sec:VQA}

Visual Question Answering (VQA) is the task of answering questions
regarding the contents of a visual scene~\citep{Malinowski2014AMA,
	antol2015vqa, kafle2017visual}. For instance, questions such as "How
many black dogs are left to the tree?" The scenarios in this task deal with multimodal data (images, questions in plain text), require structure-to-structure mapping, and symbolic reasoning has to be applied to deal with relations described in both images and text.

VQA has experienced tremendous progress in recent years due to novel
neuro-symbolic methods (see section~\ref{sub:neuro-symbolic}) and well designed
benchmarks (see Section~\ref{sec:vqa benchmarks}).

The CLEVR data set (see Section~\ref{sec:vqa benchmarks}) was especially influential in the development of better VQA methods. Early VQA algorithms achieved very modest
accuracies~\citep{johnson2017clevr} on CLEVR but rapid progress was
made. Current methods can answer very complicated questions such as
"There is an object that is both on the left side of the brown metal
block and in front of the large purple shiny ball; how big is it?"
with near perfect and super-human accuracy~\citep{yi2018neural}.
However, \cite{kafle2017analysis} point out that CLEVR is
specifically designed for compositional language approaches and
requires demanding language reasoning, but only limited visual
understanding due to the simple objects.

Beyond images, questions and answers to the CLEVR data set also provide
so called "ground-truth programs" for each sample, which define
functional programs that answer the sample question. Early Neural Module Network (NMN)
approaches for VQA required all ground-truth programs for training
\citep{hu2017learning, Suarez2018DDRprogAC}, while later methods used
less and less ground-truth programs~\citep{johnson2017inferring}. The
VQA algorithm by \cite{yi2018neural} uses only a small subset, and the
recent \emph{StackNMN}~\citep{hu2018explainable} or the
\emph{Neuro-symbolic concept learner}~\citep{mao2019neuro} learn
without any ground-truth programs at all, using images, questions and
answers alone.  Finally, \cite{shi2019explainable} demonstrated that
the \emph{Neural module networks} framework can achieve perfect
accuracies on the CLEVR benchmark provided the objects and their
relations (scene graph) are perfectly extracted from the scene.

\subsection{Compositional and hierarchical representations}
\label{sec:hierarchical_representations}

Compositional and hierarchical representation are often used to solve
some of the most challenging tasks, including visual question answering,
visual grounding, and compositional learning and reasoning. Most of
the existing work in the field are focused on tasks that make use of multi-modal data sources (images and text), and we describe those methods in section~\ref{sec:hierarchical_multi_modal}. Tree representations, and in particular Tree-LSTMs are also a popular trend to represent and generate compositional knowledge, and we cover it in section~\ref{sec:tree-lstm}. Next, we introduce hyperbolic embeddings in section~\ref{sec:hyperbolic_embedding}, where we describe methods that use hyperbolic vector spaces for better capturing hierarchical dependencies. Finally, another area where compositional and hierarchical representations are required is the automatic learning of knowledge graphs, which we describe in section~\ref{sec:knowledge graphs}.

\subsubsection{Hierarchical representations in multi-modal data}
\label{sec:hierarchical_multi_modal}

In order to deal with multimodal data, \cite{lu2016hierarchical}
proposed a co-attention mechanism for performing visual question
answering by attending to both textual descriptions and subareas in
images. This approach makes it possible to answer compositional
questions such as "what is the man holding a snowboard on top of a
snow covered?", "how many snowboarders in formation in the snow, four
is sitting?", etc.

\cite{agrawal2017c} proposed \emph{compositional-VQA} (C-VQA) which is
created by rearranging the VQAv1 dataset (see  Section~\ref{sec:vqa benchmarks}) so that QA pairs in the C-VQA
test data set are not present in the C-VQA training data set, but most
concepts constituting the QA pairs in the test data are present in
the training data. The idea is to ensure that the question-answer (QA)
pairs in the C-VQA test data are compositionally novel with respect
to those in the C-VQA training data. They evaluate existing VQA models under this new setting, and show that the performance degrade considerably.

\cite{hu2017modeling} propose \emph{compositional modular networks}
(CMN) \footnote{http://ronghanghu.com/cmn} to localize a referential
expression by grounding the components in the expressions and
exploiting their interactions, in an end-to-end manner. The model
performs the task of grounding in a three-step procedure: (a) an
expression is parsed into subject, relationship and object with
attention for language representation; (b) the subject or object is
matched with each image region with unary score; and (c) the
relationships and region pairs are matched with a pairwise score.  To
perform the same task, \cite{choi2018learning}
%argue that existing
%solutions rely on the association between the holistic language
%features and visual features, neglecting the compositional %reasoning
%implied in the language. They 
propose the Recursive Grounding Tree
(Rvg-Tree), which is inspired by the intuition that any language
expression can be recursively decomposed into two constituent parts,
and the grounding confidence score can be recursively accumulated by
calculating their grounding scores as returned by the sub-trees.

\cite{purushwalkam2019task} consider that in order to perform
compositional reasoning, it is crucial to capture the intricate
interactions between the \textit{image}, the \textit{object} and the
\textit{attribute}; while existing research only captures the
contextual relationship between objects and attributes. Assuming the
original feature space of images is rich enough, inference entails
matching image features to an embedding vector of object-attribute
pairs. They propose the Task-driven Modular Networks (TMN) to perform
the task of compositional learning.

\subsubsection{Tree-LSTMs and tree structures}
\label{sec:tree-lstm}

%As explained in Section~\ref{sec:understanding}, compositionality is
%an important aspect of understanding, and hierarchical tree structures
%have been widely used for this purpose.  

A popular hierarchical structure that has been widely used for compositionality are tree representations. A number of neural approaches
have been proposed to handle tree structures in different ways: as
input, for representation, and for generation. One of the earliest
models, recursive neural networks, was developed by
\cite{socher2013recursive}. In this model, sequential data
(e.g. natural language sentences) together with tree-structures
(e.g. the constituency parse tree of the sentence) are fed to the
neural network, which represents terminal nodes with word embeddings
and non-terminal nodes with the \textit{concatenation} of all their
child nodes. This tree-structured representation is applied to
sentiment classification. With the new representation, the system is
capable of performing sentiment classification in a more fine-grained
level by classifying non-root nodes instead of only root nodes of the
tree.

%In the latter application, the representation of the root nodes
%is compared with different similarity scores for calculating the
%semantic similarity between two sentences.

The tree-LSTM neural network \citep{tai2015improved} took this work
one step further by replacing the \textit{concatenation} operation
with more sophisticated gates. Its implementation is different from
the generic LSTM, which calculates the hidden states at each step by
using input, output, and forget gates on hidden states from previous
steps and input data of the current step. Tree-LSTMs apply the input,
output and forget gates for each non-terminal node based on hidden
states of all child nodes, and the input of the current non-terminal
node. \cite{tai2015improved} presented two implementations of
Tree-LSTM: Child-Sum TreeLSTM and N-ary Tree-LSTMs.

The vanilla Tree-LSTM assumes input from both texts and their
corresponding tree-structures. \cite{choi2018learning} propose the
Gumbel Tree-LSTM which learns to compose task-specific tree structures
only from plain text data. The model uses a Straight-Through
Gumbel-Softmax estimator to decide on the parent node among candidates
dynamically, and to calculate gradients of the discrete decision.

There are multiple versions of tree-structured LSTMs, however
\cite{Havrylov2019Cooperative} demonstrate that the trees do not
resemble any semantic or syntactic formalism. They propose latent-LSTM
to model the tree learning task as an optimization problem where the
search space of possible tree structures are explored. Learning new
concepts via structure space searching is not uncommon;
\cite{lake2018emergence} proposed a new computational model to
discover new structure organizations by using a broad hypothesis space
with a preference for sparse connectivity. Different from
probabilistic inference, where a predefined set of structure forms
(e.g. tree, ring, chain, grid) are provided, here they explore more
general structures. \cite{demeter2019just} presented the Neural-Symbolic Language Model
(NSLM) which consists of two components: a hierarchical Neural network
language model (NNLM) which assigns a probability to each class, $C$,
and a micro-model that allocates this probability over the words in
$V_C$. They specifically defined a micro-model to incorporate logic
for words referring to numbers. Note that the logic aspect is modelled
at the leaf node level, i.e. for individual classes of vocabularies.

Finally, apart from representation, some also introduce methodologies
for generating tree structures. \cite{corro2018differentiable}
proposed the tree-structured variational auto-encoder (VAE) for such
purposes. A generic VAE \citep{doersch2016tutorial} is composed of an
encoder, which extracts the hidden representation, $z$, of the input
data, $x$, and a decoder which reconstructs the input data by performing a normalized sampling $N(\mu,
\sigma)$ from the latent representation, $z$, (note that the parameters $\mu$ and
$\sigma$ are parameters to be learned). In the tree-structured VAE,
the encoder is replaced with a level-by-level representation
extraction process that \textit{merges} the representations of
terminal nodes into the representation of non-terminal nodes in higher
levels of the tree. On the other hand, the decoder is replaced with a
generative process that \textit{splits} the extracted representation
from the encoder into representations of non-terminal nodes and
finally terminal nodes. The values of terminal nodes are determined by
their representation via \textit{argmax}. In both the encoder and the
decoder, the \textit{merge} and \textit{split} operations are
implemented using multi-layered perceptrons (MLP's). Using
tree-structured VAE's \cite{jin2018junction} propose the automatic 
generation of molecular graphs.

\subsubsection{Hyperbolic embeddings}
\label{sec:hyperbolic_embedding}

As seen above, learning embeddings
(e.g. Word2Vec~\citep{mikolov2013distributed},
Glove~\citep{pennington2014glove}) of symbolic data such as text,
image parts or graphs are a common approach to capture the semantics
of entities and enable similarity measures between them. While data
often exhibits latent hierarchical structures, most methods generate
embeddings in Euclidean vector space, which do not capture
hierarchical dependencies adequately. Other embeddings such as
\emph{linear relational embeddings} \citep{paccanaro2001learning},
\emph{holographic embeddings} \citep{nickel2016holographic},
\emph{complex embeddings} \citep{trouillon2016complex} and \emph{deep
	recursive network embeddings} \citep{zhu2020deep} have been proposed
to address this issue. Here we focus on \emph{hyperbolic embeddings},
which have demonstrated better model generalisation, and more
interpretable representations for data with underlying hierarchical
structure.

\cite{nickel2017poincare} introduce an algorithm to learn a
hyperbolic embedding in an $n$-dimensional Poincar{\'e} ball and show
improved representation of WordNet \citep{miller1995wordnet} noun
hierarchies in this space compared to the traditional Euclidean
embedding. Further improvements of the method were achieved by \cite{ganea2018hyperbolic}
and \cite{de2018representation}, where the latter achieved a nearly perfect reconstruction of
WordNet hypernym graphs.

Similarly, \cite{chamberlain2017neural} employ Poincar{\'e}
embeddings to achieve improved representation on Zachary's karate
club \citep{zachary1977information} and other small-scale
benchmarks. \cite{mathieu2019continuous} modify Variational Auto
Encoders \citep{kingma2014adam} to embed the latent space in a
Poincar{\'e} ball and demonstrate better generation of MNIST
handwritten digits \citep{lecun-mnisthandwrittendigit-2010}.

While Poincar{\'e} embeddings are easy to implement, the application
for deep hierarchies is problematic due to the limited numerical
accuracy on digital computers. For instance, on the WordNet graph
approximately 500 bits of precision are needed to store values from
the combinatorial embedding
\citep{de2018representation}. \cite{yu2019numerically} address this
problem by proposing a tiling-based model for hyperbolic embeddings.
Noteworthy in this context is also the paper by
\cite{nickel2018learning}, where they revisit their earlier work on
Poincar{\'e} and find that learning embeddings in the Lorentz model is
substantially more efficient than in the Poincar{\'e}-ball model.

Considering that a crucial part of understanding is to extract
structured representations of text, the work by
\cite{le2019inferring} on inferring concept hierarchies from text
corpora via hyperbolic embeddings is especially relevant. Another
fundamental capability is structure-to-structure mapping, which has
been addressed by \cite{alvarez2019unsupervised}, using unsupervised
hierarchy matching in hyperbolic spaces.  Finally the works by
\cite{suzuki2019hyperbolic} and \cite{nagano2019differentiable} are
potentially useful to learn the assembly of neural network modules
with a differentiable method instead of reinforcement learning.

%\citep{kacmajor2020semantic}: Semantic Relatedness and Taxonomic Word Embeddings
%\citep{hajri2019learning}: Learning graph-structured data using Poincar$\backslash$'e embeddings and Riemannian K-means algorithms
%\citep{kolyvakis2019hyperkg}: HyperKG: Hyperbolic Knowledge Graph Embeddings for Knowledge Base Completion
%\citep{suzuki2019hyperbolic} : Hyperbolic Ordinal Embedding

\subsubsection{Knowledge graph learning}
\label{sec:knowledge graphs}

%% Antonio
A system with understanding capabilities needs to access structured information to reason on.
An example of such resources are knowledge graphs that are defined by entities as nodes and relations of different types as edges~\citep{wang2014knowledge}.
They are quite popular and have been used in AI in many domains, from linguistic tools such as WordNet~\citep{fellbaum2012wordnet} to domain specific resources such as OBO Foundry~\citep{smith2007obo} ontologies.
These resources are expensive to build manually, thus methods have been developed to complement these resources or exploit the existing knowledge using automatic means~\citep{paulheim2017knowledge,Dash2019HypernymDU}.

There is an existing body of literature about learning knowledge from unstructured sources such as text.
NLP methods have been extensively studied within information extraction and retrieval~\citep{mao2019bootstrapping}.
State-of-the-art deep learning based methods might be considered to learn additional information and special requirements are needed to validate the newly extracted information and to integrate it into an existing knowledge graph.

Research has been devoted as well to extend the knowledge that is already in an existing knowledge graph.
Examples of such methods include link prediction methods~\citep{zhang2018link, nickel2015review}, which specially focus in predicting whether two nodes in a graph should be linked.
An additional body of research considers existing knowledge to infer additional facts using tensors~\citep{socher2013reasoning}.
Tensor based methods have problems with large knowledge graphs or focus on a single relation~\citep{cai2018comprehensive}; graph embedding methods solve both issues~\citep{wang2017knowledge, yang2014embedding}.
Existing methods rely on positive information, and a uniform distribution is typically considered to solve this problem which is not ideal. Adversarial methods have been applied~\citep{cai2017kbgan} to alleviate this problem.

\subsection{Scene and document understanding}
\label{sec:scene_graph}

Understanding an image scene requires comprehension of the objects it contains, and the relationships between them. Objects themselves are typically composed of parts, and a complete description of a scene requires a hierarchical, multi-level representation, such as a scene
graph, where nodes represent objects or parts, and link represents relationships. Similarly, document structure understanding aims at identifying physical elements in a document layout (images, tables, lists, formulas, text, etc.), and representing their relations as a tree structure.

In the following we firstly discuss a selection of state-of-the-art
work in instance segmentation and object detection (Section~\ref{sec:object detection}). Those methods do
not extract scene graphs, but provide a basis by extracting the
objects in a scene. Next, we will discuss inverse graphics and graphic
generation approaches that attempt to learn programs to (re)create an
image (Section~\ref{sec:inverse graphics}). The learned program often can be interpreted as a structured
and hierarchical scene representation. We then discuss methods
with a focus on image decomposition and scene representation in Section~\ref{sec:image decomposition}. Finally, we present techniques for document understanding in Section~\ref{sec:document understanding}.

\subsubsection{Object detection and instance segmentation}
\label{sec:object detection}

Object detection and instance segmentation are fast evolving fields
and we point to a recent review paper by \cite{zhao2019object} for an
overview. Here we focus on a few, very recent, state-of-the-art
methods that could serve as the basis of a scene graph extractor.
\cite{Lee2019CenterMaskR} introduce \emph{CenterMask}, a real-time
anchor-free instance segmentation method that outperforms all previous
state-of-the-art models at a much faster speed by adding a spatial
attention-guided mask branch to the \emph{FCOS} \citep{tian2019fcos}
object detector. Another method for instance segmentation by
\cite{wang2019solo} is worth mentioning, since it is conceptually
very simple with an accuracy on par with Mask R-CNN \citep{he2017mask}
on the COCO data set \citep{lin2014microsoft}. Finally a very recent
improvement on Mask R-CNN with respect to speed and accuracy is
\emph{BlendMask} by \cite{chen2020blendmask} that also builds on the
\emph{FCOS} \citep{tian2019fcos} object detector.

All the methods above are supervised approaches that are fast and
accurate but require large amounts of labeled data and struggle with
object occlusion. Generative, probabilistic models are an alternative
approach that attempts to address these issues. An early example is
\emph{DRAW} \citep{gregor2015draw}, a recurrent neural network to
recognize, localize and generate MNIST digits (among other tasks). A
faster method (\emph{(D)AIR}) was proposed by \cite{eslami2016attend}
and recently has been further improved by \cite{Stelzner2019FasterAW}
and \cite{yuan2019generative}.

\subsubsection{Inverse graphics}
\label{sec:inverse graphics}

A different approach to model or understand an image is \emph{Inverse
	graphics} \citep{baumgart1974geometric}. Here the task is to learn a
mapping between the content of an input image and a probabilistic or
symbolic scene description (e.g. a graphical DSL such as
HTML). Fundamentally this is achieved by reconstructing the input
image from the scene description and using the differences between the
original image and its reconstruction as a training signal (e.g. via
reinforcement learning or MCMC). This approach is of special interest,
since it typically results in a hierarchically structured and easy to
interpret representation of a scene that generalizes well.

\cite{kulkarni2015picture} build upon the \emph{differentiable
	Renderer} \citep{loper2014opendr} to create "Picture", a
probabilistic programming framework to construct generative vision
models for different inverse graphics applications. They demonstrate
its usefulness on 3D face analysis, 3D human pose estimation, and 3D
object reconstruction.

Originally inverse graphics relied largely on probabilistic scene
representations. Later work is including symbolic representations and
a good example is the work by \cite{wu2017neural}, where an
encoder-decoder architecture with a discrete renderer is used to learn
image representations in an XML-flavored scene description
language. Similarly, \cite{Zhu2018AutomaticGP} employ a
encoder-decoder architecture based on hierarchical LSTM's to map
screenshots from Graphical User Interfaces (GUI) to a DSL that describes
GUI's, and \cite{NIPS2018_7845} learn to infer graphics programs in
\LaTeX\ from hand-drawn sketches, using a combination of deep neural
networks and stochastic search. Finally, a neuro-symbolic system for
inverse graphics based on a capsule network architecture
\citep{hinton2011transforming} was proposed by
\cite{kissner2019neural}.
%\citep{young2019learning} : Learning Neurosymbolic Generative Models via Program Synthesis

\subsubsection{Image decomposition}
\label{sec:image decomposition}

The third approach, beyond inverse graphics and object detection, to
understand or structure image content is to decompose the
image into parts and objects, and to construct some form of scene
graph.  For instance, \cite{burgess2019monet} introduce the
Multi-ObjectNetwork (MONet), in which a Variational Autoencoder (VAE)
together with a recurrent network is trained in an unsupervised manner
to decompose 3D scenes such as CLEVR \citep{johnson2017clevr} into
objects and background. \cite{charakorn2020explicit} also employ VAE's
but focus on disentangled representations of scene properties. Recent
advances along this line of unsupervised scene-mixture models that
allow to separate objects from background are: IODINE
\citep{greff2019multi}, GENESIS \citep{engelcke2019genesis} and SPACE
\citep{lin2020space}.

\cite{Deng2019GenerativeHM} go one step further and propose a
generative model (RICH) that allows decomposing the objects and parts
of a scene in a tree structure. Correctly recognizing occluded objects
in a scene is a common problem in scene representation that
specifically has been addressed by \cite{Yuan2019GenerativeMO}.

%\citep{Wang2019DeepHR} :Deep High-Resolution Representation Learning for Visual Recognition
%\citep{kissner2019adding} : Adding Intuitive Physics to Neural-Symbolic Capsules Using Interaction Networks
%\citep{ha2017neural} : A neural representation of sketch drawings

\subsubsection{Document understanding}
\label{sec:document understanding}
%%% Antonio: should we keep this section?

A large portion of existing data is available in unstructured document
formats such as PDF (portable document format), with over 2.7 trillion
documents available in this format. Machines that can understand
documents will improve the access to information that is otherwise
difficult to process.

Document layout and structure understanding aims at parsing
unstructured documents (e.g., PDF, scanned image) into machine
readable format (e.g., XML, JSON) for down-stream
applications. Document layout analysis identifies physical elements
(image, table, list, formula, text, title, etc) in a document, without
logical relations between the elements. Document structure analysis
aims at representing documents as a tree structure to encode the logical
relations.

Deep neural networks have been used to understand the physical layout
and logical relations in documents. Convolutional neural
networks~\citep{hao2016table,chen2017convolutional},
fully-convolutional neural
networks~\citep{he2017multi,kavasidis2019saliency}, region-based
convolutional neural
networks~\citep{schreiber2017deepdesrt,gilani2017table,staar2018corpus,zhong2019publaynet},
and graph neural networks~\citep{qasim2019rethinking,renton2019graph}
have all been exploited to parse the physical layout of documents or
to detect elements of interest (e.g., tables). Encoder-decoder
networks~\citep{zhong2019image} and visual relation extraction
networks~\citep{nguyen2019multi} have been adopted to infer the logical
structure of documents. Layout information is also integrated in
language models for NLP tasks that take unstructured documents as
input~\citep{xu2019layoutlm}.

%Current document understanding works are limited to high-level parsing of the
%documents (e.g., table detection, layout analysis, which section a paragraph
%belongs to). None of the existing frameworks support the fine-grained structured
%knowledge representation learning from a document, which is a critical part of our ultimate vision (user manual understanding). We aim to develop an AI system
%that can create a structured knowledge representation of a document by
%identifying and linking entities mentioned in various parts of the document. The
%structured knowledge representation will then engine reasoning and answering
%complex questions.

\subsection{Neuro-symbolic computing}
\label{sub:neuro-symbolic}

\subsubsection{Neuro-symbolic reasoning}
\label{sub:symbolic reasoning}

Neural networks (NN) are well suited to learn from data, and have done exceptionally well for various data modalities including structured, natural language and imaging data \citep{lecun2015deep}. Despite the ability of NN in achieving state-of-the-art accuracies \citep{Tan2019EfficientNet}, they fail at gaining a fundamental understanding of the data and rather focus on the distributions and statistical relations in the data. Another disadvantage of NN is that they often require large volumes of data in order to achieve these state-of-the-art accuracies. On the other side of the spectrum there is logical reasoning, which uses a set of facts and rules to make inferences. This allows direct interpretability of the model, something neural networks are not able to do. One of the downsides to logical reasoning is that they need symbolic knowledge, which needs to be extracted from data first before being applied in the reasoning. Logical neural networks (LNN) aim at combining these two approaches, that is to leverage the ability of neural networks to represent and learn from data and the ability of formal logic to perform reasoning on what has been learned by NN \citep{Garcez2019neuralsymbolic}. 

Although the research area of combining NN and logical reasoning has received a lot of interest in recent years \citep{Garcez2019neuralsymbolic}\citep{manhaeve2018deepproblog}\citep{serafini2016logic}\citep{Raedt2019NeuroSymbolicN}\citep{dong2019neural}\citep{Riegel2020LogicalNeuralNetworks}, the idea of combining machine learning, and in specific neural networks, and logical reasoning has been described in early works of \citep{chan1993neural}, \citep{Tong-Seng1995Utilizing}, \citep{Muggleton1991Inductive} dating back to the early 90's. \cite{Garcez2019neuralsymbolic} wrote a review paper on some strategies and methodologies employed for neuro-symbolic approaches and outlines the key characteristics of such a system, namely:

\begin{enumerate}
	\item Knowledge representation \label{item:Knowledge representation}
	\item Learning \label{item:Learning}
	\item Reasoning \label{item:Reasoning}
	\item Explainability (Interpretability) \label{item:Explainability}
\end{enumerate}

\ref{item:Knowledge representation} Knowledge representation: How will the available knowledge be represented? Are there explicit rules that can be represented by propositional of first order logic which would allow logical reasoning using logical programs. 

\ref{item:Learning} Learning: \cite{Garcez2019neuralsymbolic} mentions two popular learning approaches, horizontal and vertical learning and describes how neural learning and logical reasoning is interfaced in each. \cite{Hu2019Harnessing} explains the horizontal learning approach, making use of a 'teacher' and 'student' network. First, the 'student' network gets trained on the available data. Once trained, this 'student' network is projected into a 'teacher' network. Secondly, the prior knowledge (rules) are added to the 'teacher' network by regularizing the network's loss function. This loss from the 'teacher' network is then used to optimize the weights of the 'student' network. Thus, this approach offers a way to distill prior knowledge into the neural network weights, and serves as a knowledge extraction method. The disadvantage of such approaches is that the rules are integrated into the final network and not available at inference time for further analysis. \cite{manhaeve2018deepproblog} explains one approach to vertical LNN learning. The first part of such a network is a neural network, such as a CNN, RNN etc. that performs low level perception. In the case of the MNIST handwritten digits \citep{lecun-mnisthandwrittendigit-2010} example, the CNN is used to extract the low level features and output a probability score (softmax) for each digit. The second part of the network is logical reasoning, which could for example be the addition of two digits, defined in the logic of ProbLog \cite{DeRaedt2007ProbLog}.The network is end-to-end differentiable and allows for uncertainty in the final answer by using probabilities.

\ref{item:Reasoning} Reasoning: Arguably one of the most desirable character traits an AI system should have. Apart from being able to learn from available data, the system should be able to reason on the learned features in order to come to a decision. This can be achieved by the explicit use of formal logic as done by \cite{manhaeve2018deepproblog}. Other examples of reasoning approaches is that by \cite{Hu2019Harnessing}, whom incorporates a confidence variable to allow for uncertainty and achieves logical reasoning where not all rules can be satisfied exactly. \cite{serafini2016logic} uses approximate satisfiability to allow for flexibility in rules being satisfied. \cite{Riegel2020LogicalNeuralNetworks} makes use of truth bounds, consisting of a lower and upper bound [L, U] which can be used to assign truth (L = U = $[\alpha, 1]$), false (L = U = $[0, 1-\alpha]$) as well as unknown (L = $[0, 1-\alpha]$, U = $[\alpha, 1]$) values to an outcome or particular node in the network. In addition to the latter, the network can also highlight cases where contradicting (L $>$ U) knowledge arise in the network.

\ref{item:Explainability} Explainability (Interpretability): In the last couple of years there has been an increasing interest in the ability to explain and interpret the models decision and to understand the reasoning behind a given result \citep{Fan2020OnInterpretability}. Perhaps an important point to outline here is the distinction between interpretability and explainability. The former is a model that provides insight into decision making on its own without the need for additional processing, as apposed to the latter that requires an additional model or processing to explain the original model. Accuracy is therefore no longer the most important criteria for model performance and is especially true in domains where models are used in life critical decision making such as health care, and where ethical decision making is critical. These model needs to be interpretable and should allow transparency and understanding of the reasoning behind a particular model outcome. Logical statements are easily interpretable and thus account for the model interpretability. However, interpretability of the model as a whole depends on how the knowledge was represented to the logical part of the model. As described above most learning strategies depend on either a horizontal or vertical approach. Thus, the part performed by the neural network could cloud full interpretability of the model.

The above methods all makes use of explicit logic in some form or another. Neural networks are used to learn features and extract symbolic representations in the images and logic is used in a subsequent reasoning step. An alternative, non-logic, approach is presented by \cite{andreas2016neural}. The author introduces \emph{Neural Module Networks} (NMN), a composition of multiple neural networks in a jointly-trained arrangement of modules for question answering. The most interesting aspect of this method is that the arrangement of the modules is dynamically inferred from the grammar and content of the question. This not only enables long-chained reasoning but also provides interpretabily with detailed text and image attention maps \citep{Lu2016HierarchicalQC}. This method offers the advantage that the domain knowledge does not need to be known upfront as is the case with most of the methods requiring exact logic. This advantage could be a disadvantage when exact knowledge is available and could help provide full model interpretability. An example of this could be in medical diagnoses, where a disease is diagnosed according to a set of guidelines. These guidelines can be encoded in logical statements and provide interpretabilty of the model.

In \citep{andreas2016learning}, the authors improved Neural module networks, and learned the arrangement of the modules jointly with the answers instead of relying on a language parser for the questions and handwritten rules for module assembly. These \emph{Dynamic Neural Module Networks} (DNMN) are more accurate and understand more complex questions. Further improvements were achieved by \cite{hu2017learning} and their \emph{End-to-End Module Networks} 
(N2NMM).

\subsubsection{Program synthesis}
\label{sec:program_synthesis}

Program synthesis is the task of learning programs (code) from examples of input and output data. This task is strongly linked to an AI that understands, since induced programs provide explainability (via symbolic rules),
enable reasoning, and also result in a structured representation of an input. 
%The modules in NMNs represent functions in a larger program and the
%problem of inferring such programs (the composition of functions)
%based on input-output examples is fundamentally a \emph{Program
%  Synthesis} task \citep{kant2018recent}. 
Early attempts utilized Genetic programming \citep{banzhaf1998genetic} to evolve programs \citep{koza1992genetic}, and interest in the field has gained new strength with the success of program-like neural network models. This section describes recent techniques for program synthesis and how they relate to an AI that understands.

\cite{Parisotto2016NeuroSymbolicPS} introduce \emph{Recursive-Reverse-Recursive Neural Network} (R3NN) to incrementally construct programs conforming to a pre-specified Domain Specific Language (DSL). R3NN is a tree structured generation model that is trained with supervision, and it was able to solve the majority of Flash-Fill benchmarks \citep{gulwani2011automating}, which consist on synthesizing programs built on regular expressions to perform the desired string transformation. However R3NN did not generalize well to larger programs nor does it seem to scale well computationally for more complex tasks. Poor generalization is a known issue in program synthesis.  \cite{Cai2017MakingNP} address this problem by augmenting neural architectures with recursion. They apply \emph{Neural Programming Architecture} (NPA) to four tasks: grade-school addition, bubble sort, topological sort, and quicksort; and they demonstrate that recursion improves accuracy.

\cite{ellis2015unsupervised} employ general-purpose symbolic solvers for Satisfiability Modulo Theories (SMT) problems \citep{de2008z3} to learn SVRT (Synthetic Visual Reasoning Test) concepts \citep{fleuret2011comparing} such as spatial relationships between objects. Training is unsupervised, but the symbolic solvers do not scale well to more complex problems and there is no soft-reasoning.  One reason why symbolic solvers do not scale well is that the search space grows exponentially. \emph{DeepCoder} \citep{balog2016deepcoder} addresses this problem by learning an embedding for the functions of a DSL to guide the (beam) search for program candidates. \cite{Zohar2018AutomaticPS} were able to further improve and accelerate \emph{DeepCoder} by learning to remove intermediate variables.

\cite{Feser2016DifferentiableFP} implemented a differentiable program interpreter utilizing a DSL inspired by functional programming, but they found, in agreement with \citep{gaunt2016terpret}, that discrete search-based techniques for program synthesis such as $\lambda^2$ \citep{feser2015synthesizing} perform better than differentiable programming approaches. Later work by \cite{Gaunt2016DifferentiablePW} extend a differentiable programming language with neural networks, enabling the combination of functions with learnable elements, in a similar fashion to Neural Network Modules \citep{andreas2016neural} with symbolic modules. Discrete search-based techniques for program synthesis are also a focus of \cite{chen2017towards}, who in addition highlight the importance of recursion. Using reinforcement learning, they demonstrate significantly improved generalization for learning context-free parsers. Similarly, improved generalization for equation verification and completion was achieved by \cite{Arabshahi2018CombiningSA} using \emph{Tree-LSTMs} \citep{tai2015improved} with weakly supervised training. Recent work by \cite{Pierrot2019LearningCN} further advances the above line of research, introducing the novel reinforcement learning algorithm \emph{AlphaNPI} with structural biases for modularity, hierarchy and recursion.

\emph{Program induction} is similar to \emph{Program Synthesis} in that a program needs to be learned from input-output examples, however, in \emph{Program induction} the program is not explicit and not part of the result \citep{kant2018recent}. In spirit, methods for Program induction tend to be closer to neural networks than to symbolic computing. For instance, architectures such as the \emph{Neural Turing Machine} \citep{graves2014neural, graves2016hybrid}, the \emph{Differential Neural Computer} \citep{graves2016hybrid,Tanneberg2019LearningAS}, the \emph{Neural programmer} \citep{hudson2019learning}, \emph{Neural programmer-interpreters} \citep{reed2015neural, Pierrot2019LearningCN}, \emph{Neural Program Lattices} \citep{Li2017NeuralPL}, the \emph{Neural State Machine} \citep{hudson2019learning}, and most recently \emph{MEMO} \citep{banino2020memo} extend neural networks with external memory, and can infer simple algorithms such as adding numbers, copying, sorting and path finding. \cite{manhaeve2018deepproblog} illustrates the use of \emph{DeepProbLog} to solve three program induction tasks and compare their results to Differentiable Forth ($\partial4$) \citep{Bosnjak2017Programming}. The first of the three examples looks at calculating the resulting digit and the carry digit given the sum of two digits and the previous carry digit. The results between DeepProbLog and $\partial4$ where the same, with both obtaining 100\% accuracy for training lengths of up to 8 digits and 64 testing digits. The second program induction problem they looked at was the bubble sort algorithm. Given a list of unsorted digits, the task is to arrange the digits in a sorted list. The program was tasked to figure out the action in each step of the bubble. DeepProbLog showed higher accuracy than $\partial4$ for input lists of length bigger than 3 and did so in a fraction of the time $\partial4$ was able to. The final program induction problem was solving algebraic word problems, where the program had to decide on the order of input digits and the mathematical operation required in each step. As with the first problem, similar results were obtained between DeepProbLog and $\partial4$, 96.5\% and 96.0\% respectively.

\subsection{Commonsense inference}
\label{sec:common_sense_inference}
%% Ella
An intelligent creature needs to know about the real world and use its knowledge effectively to be able to act sensibly in the world. The knowledge of a schoolchild about the world and the methods for making obvious inferences from this knowledge are called common sense. Commonsense knowledge, such as knowing that
\textit{“bumping into people annoys them”} or \textit{“rain makes the road slippery”}, helps humans navigate
everyday situations seamlessly~\citep{apperly2010mindreaders}. This type of knowledge and reasoning plays a crucial role in all aspects of artificial intelligence, from language understanding to computer vision and robotics.~\citep{davis2014representations}.

\paragraph{Commonsense inference in NLP}

A successful linguistic communication relies on a shared experience of the world, and it is this shared experience that makes utterances meaningful. Despite the incredible effectiveness of language processing models trained on text alone, today’s best systems still make mistakes that arise from a failure to relate language to the physical world it describes and to the social interactions it facilitates~\citep{bisk2020experience}. One important difference between human and machine text understanding lies in the fact that humans can access commonsense knowledge while processing text, which helps them to draw inferences about facts that are not mentioned in a text, but that are assumed to be common ground. However, for a computer system, inferring such unmentioned facts is a non-trivial challenge~\citep{ostermann2019commonsense}.

In recent years, NLP community has introduced multiple exploratory research directions into automated commonsense understanding~\citep{sap2020introductory}. Recent efforts to acquire and represent this knowledge have resulted in large knowledge graphs, acquired through extractive methods~\citep{speer2016conceptnet} or crowdsourcing~\citep{sap2019atomic}. Moreover, reasoning capabilities have been integrated into downstream NLP tasks, in order to develop smarter dialogue~\citep{zhou2018commonsense} and question answering systems~\citep{xiong2019improving}.

Recent large pretrained language models~\citep{devlin2018bert, liu2019roberta, brown2020language},
however, have significantly improved human-like understanding capabilities of machines. Hence, machines sould be able to model commonsense through symbolic integrations. Given the large number of NLP applications which are designed to require commonsense reasoning, some efforts infer commonsense knowledge from structured KBs as additional inputs to a neural network in generation~\citep{guan2019story}, dialogue~\citep{zhou2018commonsense}, question answering~\citep{mihaylov2018knowledgeable, bauer2018commonsense, lin2019kagnet, weissenborn2017dynamic, musa2018answering}, and classification~\citep{chen2019incorporating, paul2019ranking, wang2019improving}. In some others,  researchers have relied on commonsense knowledge aggregated from corpus statistics exploited from unstructured text~\citep{tandon2018reasoning, lin2017reasoning, li2018multi, banerjee2019careful}. Recently,  instead of using relevant commonsense as an additional input to neural networks, commonsense knowledge has been encoded into the parameters of neural networks through pretraining on relevant knowledge bases~\citep{zhong2019improving} or explanations~\citep{rajani2019explain}, or by using multi-task objectives with commonsense relation prediction~\citep{xia2019incorporating}.

Commonsense is indirectly evaluated by assessing the performance of higher-level tasks over conventional Natural Language Understanding (NLU) data sets. Direct measurement of capabilities for assessing and explaining commonsense can shed light upon the ability to represent semantic knowledge, and symbolic reasoning with uncertainty. A detailed account of challenges with commonsense reasoning is provided by~\cite{davis2015commonsense}, which spans difficulties in understanding and formulating commonsense knowledge for specific or general domains to complexities in various forms of reasoning and their integration for problem solving~\citep{storks2019commonsense}. 

Common sense inference in NLP is normally evaluated by reading comprehension tasks that require question-answering about a text by inferring information that is common knowledge but not necessarily present in the text. \citep{ostermann2019commonsense} evaluated two commonsense inference tasks on everyday narratives (task1) and on news articles (task2), based on two data sets: MCScript2.0~\citep{ostermann2019mcscript2} and ReCoRD~\citep{zhang2018record}. These two tasks were also released as challenges in the SemEval 2018 shared task 11, and in the COIN (COmmonsense INference in Natural Language Processing) workshop at EMNLP-IJCNLP 2019. Three baseline models: logistic regression~\citep{merkhofer2018mitre}, attentive reader~\citep{hermann2015teaching} and three-way attentive network~\citep{wang2018yuanfudao}; and five baseline models: BERT~\citep{devlin2018bert}, KT-NET~\citep{yang2019enhancing}, stochastic  answer network (SAN)~\citep{liu2017stochastic}, the DocQA~\citep{clark2017simple} and random guess, were presented for task1 and task2, respectively. Systems from five teams during the COIN workshop were evaluated in~\citep{ostermann2019commonsense} among which a best accuracy of 90.6\% and a best F1-score of 83.7\% were achieved for task1 and task2, respectively. Even with the best-performing transformer-based method, machine performances are 7\% and 8\% lower than human performance in the two tasks. Also, this particular field is in demand of better data sets that make it harder to benefit from redundancy in the training data or large-scale pre-training on similar domains~\citep{ostermann2019commonsense}. 

\paragraph{Visual commonsense inference}

Visual understanding goes well beyond object recognition. Understanding the world beyond the pixels is very straightforward for humans. However, this task is still difficult for today’s vision systems, which require higher-order cognition and commonsense reasoning about the world. Various tasks have been introduced
for joint understanding of visual information and language, such as image captioning~\citep{chen2015microsoft, vinyals2015show, sharma2018conceptual}, visual question answering~\citep{antol2015vqa, johnson2017clevr, marino2019ok} and referring expressions~\citep{kazemzadeh2014referitgame, plummer2015flickr30k, mao2016generation}. There is also a recent body of work addressing representation learning using vision and language cues~\citep{tan2019lxmert, lu2019vilbert, su2019vl}. However, these works fall short of understanding the dynamic situation captured in the image, which is the main motivation of visual commonsense inference. 

Visual understanding requires seamless integration between \textit{recognition} and \textit{cognition}: beyond recognition-level perception (e.g. detecting objects and their attributes), one must perform cognition-level reasoning (e.g. inferring the likely intents, goals, and social dynamics of people)~\citep{davis2015commonsense}.  
State-of-the-art vision systems can reliably perform recognition-level image understanding, but still struggle with complex inferences. In the contect of visual question answering, some work focuses on commonsense phenomena, such as ‘what if’ and ‘why’ questions~\citep{pirsiavash2014inferring, wagner2018answering}. However, the space of commonsense inferences is often limited by
the underlying dataset chosen (synthetic~\citep{wagner2018answering} or COCO~\citep{pirsiavash2014inferring} scenes). In~\citep{zellers2019recognition}, commonsense questions are asked in the context of rich images from movies. Some visual commonsense inference work~\citep{mottaghi2016happens, ye2018interpretable} involves reasoning about commonsense phenomena, such as physics. Some involves commonsense reasoning about social interactions~\citep{alahi2016social, chuang2018learning, gupta2018social, vicol2018moviegraphs}, while some others involve procedure understanding~\citep{alayrac2016unsupervised,zhou2017towards}. Predicting what might happen next in a video is also studied in~\citep{singh2016krishnacam,ehsani2018let,zhou2015temporal, vondrick2016anticipating, felsen2017will, rhinehart2017first,yoshikawa2018stair}.

To move toward incorporating commonsense knowledge in the context of visual understanding, \cite{vedantam2015learning} proposed an approach where human-generated abstract scenes made from clipart is used to learn common sense, but not on real images. Inferring the motivation behind the actions of people from images is also explored by~\citep{pirsiavash2014inferring}. Visual Commonsense Reasoning (VCR)~\citep{zellers2019recognition} tests if the model can answer questions with rationale using commonsense knowledge. Given a challenging question about an image, a machine must answer correctly and then provide a rationale to justify its answer. To move towards cognition-level understanding, \cite{zellers2019recognition} porposed a new reasoning engine, Recognition to Cognition Networks (R2C), that models the necessary layered inferences for grounding, contextualization, and reasoning. However R2C helps narrowing the gap between humans and machines, the challenge is still far from solved. While this work includes rich visual common sense information, their question answering setup makes it difficult to have models to generate commonsense inferences~\citep{parkvisualcomet}. 

ATOMIC~\citep{sap2019atomic} on the other hand provides a commonsense knowledge graph containing if-then inferential textual descriptions in generative setting; however, it relies on generic, textual events and does not consider visually contextualized information. In further attempt, \cite{parkvisualcomet} proposed an approach by extending \cite{zellers2019recognition} and \cite{sap2019atomic} for general visual commonsense to build a large scale repository of visual commonsense graphs and models that can explicitly generate commonsense inferences for given images. This repository consists of over 1.4 million textual descriptions of visual commonsense inferences carefully annotated over a diverse set of 59,000 images, each paired with short video summaries of before and after.

From another perspective, there is also a large body of work on future prediction in different contexts such as future frame generation~\citep{ranzato2014video, srivastava2015unsupervised, xue2016visual, vondrick2016generating, mathieu2015deep, villegas2019high, castrejon2019improved}, prediction of the trajectories of people and objects~\citep{walker2014patch, alahi2016social, mottaghi2016happens}, predicting human pose in future frames~\citep{fragkiadaki2015recurrent, walker2017pose, chao2017forecasting} and semantic future action recognition~\citep{lan2014hierarchical, zhou2015temporal, sun2019relational}.

%% Antonio - Not sure if we should propose a pipeline
%%\section{Proposal of an AI pipeline that understands}

\section{Discussion}
\label{sec:discussion}

%%% Summarize what is available as tasks and methods
In previous sections, we have identified relevant characteristics for an AI that understand. These capabilities are present in benchmarks and research streams that advance the current state of the art in understanding capabilities of AI systems.
In this section, we summarize the different benchmarks and streams and present several capabilities that we considered relevant for an AI that understands.
Then, we categorize the benchmarks using those capabilities and discuss the different research streams.

\subsection{AI that understands capabilities}
\label{sub: capabilities}

Following the discussion and requirements from an AI that understands defined in previous sections, we have developed the following capabilities as relevant to an AI that understands:

%%% Define overall characteristics and reference the table
\begin{itemize}
	
	\item{Capability 1}: hierarchical and compositional knowledge representation, e.g. are the different components of an object be represented? Can we for instance compose several objects to derive a different one? Being able of decomposing objects supports better generalisation to previously unseen situations by the AI system.
	
	\item{Capability 2}: multi-modal structure-to-structure mapping. As we have seen, information is available in several modalities including text and images. In order to combine information from different modalities, a mapping from these modalities to a structured knowledge representation provides the means for understanding information from several modalities.
	
	\item{Capability 3}: integration of symbolic and non-symbolic knowledge. This is required to model information that does not have a symbolic representation with existing symbolic knowledge.
	
	\item{Capability 4}: symbolic reasoning with uncertainties. Uncertainty about the world might be present even with fully symbolic representations. Uncertainties might be due to incompleteness, inconsistencies or noise, or motivated by world changes. 
	
\end{itemize}

%%% Define characteristics of the benchmarks
\subsection{Benchmarks and understanding}
\label{sub: benchmarks discussion}

If we revisit the benchmarks using the set of capabilities above, we obtain Table~\ref{tab:benchmarks}.
One first obvious characteristic of most benchmarks is that they deal with textual or image modalities in tasks such as object detection or question answering.
As presented in Table~\ref{tab:benchmarks}, the benchmarks covering a broader set of understanding capabilities are in visual question answering category, while benchmarks purely based on images or text focus on modelling specific tasks, such as object detection, that can be considered for understanding building blocks.
Visual question answering considers a broad set of capabilities including composition of objects, their relation to each other and combination of multimodal data from text and images.
%%% Consider showing examples
The CLEVR data set supports as well symbolic reasoning with uncertainties about the world but there is no compositionality or hierarchical composition of the objects in the scene, thus the object detection methods might just focus on identifying shapes and colors and do not reason on more complex objects.
A recent benchmark named Math SemEval~\cite{hopkins-etal-2019-semeval} requires reasoning about compositionality or hierarchical composition of the data in the benchmark, thus analysis of images, text and the need to combine everything to reason about the answer.

%%% Add examples from commonsense inference benchmarks
Considering the commonsense benchmarks, there exist a large number of benchmarks focused on question answering, mostly based on natural language.
Some of these benchmarks are approached using natural language processing methods using pre-trained models on large corpora.
In some examples such as ATOMIC, a knowledge graph is provided to reason on.

%%% Some summary of what should be done next
Ideally, an AI that understands would meet the requirements for the four capabilities. We have observed that the current benchmark development is progressing in the right direction but it might be possible to expand some of the existing ones with additional capabilities.
For instance, the CLEVR benchmark could be extended to deal with a composition of objects, possibly starting with simple geometric objects and then increase the complexity into more complex objects obtained as a composition of this simpler ones.

%%% 
A recent benchmark, the Abstraction and Reasoning Challenge (ARC)\footnote{\url{https://www.kaggle.com/c/abstraction-and-reasoning-challenge}}~\cite{chollet2019measure}, has been defined to predict changes in an abstract block world in which reasoning would be used to predict the future state of the block world abstracting from small number of examples.

\begin{table}
	\begin{small}
		\begin{tabular}{c|c|c|c|c}
			Benchmarks&CAP1&CAP2&CAP3&CAP4\\ \hline
			Image analytics&&&&\\ \hline
			ImageNet~\cite{deng2009imagenet}&\xmark&\xmark&\xmark&\xmark\\
			COCO~\cite{lin2014microsoft}&\xmark&\xmark&\xmark&\xmark\\
			SHAPE benchmark~\cite{shilane2004princeton}&\xmark&\xmark&\xmark&\xmark\\
			\hline
			Natural language processing&&&&\\\hline
			SQUAD~\cite{rajpurkar2016squad}&\xmark&\xmark&\xmark&\xmark\\
			GLUE~\cite{wang2018glue}&\xmark&\xmark&\xmark&\xmark\\
			BioASQ~\cite{tsatsaronis2015overview}&\xmark&\xmark&\xmark&\xmark\\
			AI2 Reasoning Challenge&\cmark&\xmark&\xmark&\xmark\\
			LAMBADA~\cite{paperno2016lambada}&\xmark&\xmark&\cmark&\xmark\\
			BREAK~\cite{Wolfson2020Break}&\cmark&\xmark&\cmark&\xmark\\
			\hline
			Visual QA&&&&\\ \hline
			CLEVR~\cite{johnson2017clevr}&\xmark&\cmark&\cmark&\cmark\\
			VQAv2~\cite{goyal2017making}&\xmark&\cmark&\cmark&\xmark\\
			Visual Genome~\cite{antol2015vqa}&\xmark&\cmark&\cmark&\xmark\\
			Math SemEval~\cite{hopkins-etal-2019-semeval}&\cmark&\cmark&\cmark&\xmark\\
			\hline
			Common sense inference&&&&\\ \hline
			TriviaQA~\cite{joshi2017triviaqa}&\xmark&\xmark&\xmark&\xmark\\
			CommonsenseQA~\cite{talmor2018commonsenseqa}&\cmark&\cmark&\xmark&\xmark\\
			NarrativeQA~\cite{kovcisky2018narrativeqa}&\xmark&\xmark&\xmark&\xmark\\
			NewsQA~\cite{trischler2016newsqa}&\xmark&\xmark&\xmark&\xmark\\
			SWAG~\cite{zellers2018swag}&\cmark&\xmark&\xmark&\xmark\\
			WikiSQL~\cite{zhong2017seq2sql}&\cmark&\xmark&\xmark&\xmark\\
			Graph-basedLearningDataset~\cite{xu2018graph2seq}&\cmark&\xmark&\xmark&\xmark\\
			HotpotQA~\cite{yang2018hotpotqa}&\xmark&\xmark&\cmark&\xmark\\
			WorldTree~\cite{jansen2018worldtree}&\xmark&\xmark&\cmark&\xmark\\
			Event2Mind~\cite{rashkin2018event2mind}&\xmark&\xmark&\xmark&\xmark\\
			Winograd-WinogradNLIschemaChallenge~\cite{mahajan2018winograd}&\xmark&\xmark&\xmark&\xmark\\
			ConceptNet~\cite{speer2017conceptnet}&\cmark&\xmark&\xmark&\xmark\\
			WebChild~\cite{tandon2017webchild}&\cmark&\xmark&\xmark&\xmark\\
			NELL~\cite{mitchell2018never}&\cmark&\xmark&\xmark&\xmark\\
			ATOMIC~\cite{sap2019atomic}&\cmark&\xmark&\xmark&\xmark\\
		\end{tabular}
	\end{small}
	\caption{CAP1: hierarchical and compositional knowledge representation, CAP2: multi-modal structure-to-structure mapping, CAP3: integrates symbolic and non-symbolic knowledge, CAP4: supports symbolic reasoning with uncertainties}
	\label{tab:benchmarks}
\end{table}

%%% Define characteristics of the research streams
\subsection{Research streams}

The capabilities defined above describes high level functionality an AI that understand should have, but in order to obtain these capabilities would require one or more lower level building blocks. These lower level building blocks could be from several research fields, or research streams as referred to in this text, combined in a single model architecture that allows the problem at hand to be solved.

The CLEVR dataset is a good a example to illustrate this concept. This is a Visual Question Answer (VQA) dataset containing a number of 3D geometrical shapes arbitrarily arranges with respect to each other. Based on questions such as "Are there an equal number of large things and metal spheres?" or "How many objects are either small cylinders or metal spheres"? the corresponding answer is inferred.  This requires identification of attributes in the image, counting of objects, comparison between objects to be made, multiple attention and logical operations. The building blocks required to solve this VQA problem includes image analytics, such as a feature extractor using CNN, text encoding using word embeddings and subsequent processing by an LSTM, the NLP component. The use of attention layers allows focus to be placed on specific areas of interest and thus creates a symbolic representation of the input data. The output layer provides probabilities of the outcome and thus reasoning with uncertainty.  This example illustrates the combination of multiple research fields or streams to enable an AI that shows understanding. As shown in Table~\ref{tab:benchmarks}, this benchmarks captured three of the four AI requirements.

The streams in section~\ref{sec:review} are considered relevant to an AI that understand and applicable to address the benchmarks presented in section~\ref{sec:benchmarks}. Future development of these streams, and others, would allow more complex benchmarks to be solved. in turn, the creation of more complex benchmark also guides and drives the developments of the respective research streams. The relative new interest and developments in neuro-symbolic computing is a particularly interesting area of research and has sparked new interest is system that can reason and offer model interpretability.

\section{Conclusion}
\label{sec:conclusion}

The paper considers the components an artificial intelligence system that understands should have. That is, a system that not only learns statistical relationships within the data, but is capable of forming a human-like understanding of the input data. This is most certainly a truly difficult problem to solve and the purpose of the paper is not to claim a general solution for solving it, but to look at several research streams and some of their latest developments. Furthermore, several benchmarks are described, which have been used to study certain characteristics of an AI that understands. The work also contributes to a growing interest in artificial intelligence systems that are interpretable and transparent; properties that are crucial in domains, such as medical diagnosis.

% -----------------------------------------------------------------------------------------------------------------------
\section*{References}

\bibliography{references}

\end{document}